\documentclass[10pt,twocolumn,letterpaper]{article}

\usepackage[pagenumbers]{cvpr}

\usepackage{microtype}

\AtBeginDocument{\flushbottom}

\AtBeginDocument{%
  \setlength{\abovedisplayskip}{5pt plus 2pt minus 2pt}%
  \setlength{\belowdisplayskip}{5pt plus 2pt minus 2pt}%
  \setlength{\abovedisplayshortskip}{3pt plus 2pt}%
  \setlength{\belowdisplayshortskip}{4pt plus 2pt minus 2pt}%
}

\definecolor{cvprblue}{rgb}{0.21,0.49,0.74}
\usepackage[breaklinks,colorlinks,allcolors=cvprblue]{hyperref}

\title{Composable Driving Style for End-to-End Autonomous Driving}

\author{%
{\fontsize{11}{13}\selectfont
Guantian Zheng\textsuperscript{1}\quad
Lidong Yu\textsuperscript{2}\quad
Tianyu Gao\textsuperscript{3}\quad
Jiashu Li\textsuperscript{3}}\\
{\fontsize{11}{13}\selectfont
Jiaxing Chen\textsuperscript{4}\quad
Wilfred Ariel Mulyawan\textsuperscript{1}\quad
Chen Lv\textsuperscript{1,\ensuremath{\dagger}}}\\[2pt]
{\fontsize{9}{11}\selectfont
\textsuperscript{1}Nanyang Technological University\quad
\textsuperscript{2}NEOLIX\quad
\textsuperscript{3}The Chinese University of Hong Kong\quad
\textsuperscript{4}Tsinghua University}%
}

\hypersetup{%
    pdftitle={Composable Driving Style for End-to-End Autonomous Driving},
    pdfauthor={Guantian Zheng, Lidong Yu, Tianyu Gao, Jiashu Li, Jiaxing Chen, Wilfred Ariel Mulyawan, Chen Lv}
}

\begin{document}
\maketitle
\begingroup
\renewcommand{\thefootnote}{\ensuremath{\dagger}}
\footnotetext[1]{Corresponding author. Contact email:
\href{mailto:gzheng004@e.ntu.edu.sg}{gzheng004@e.ntu.edu.sg}.}
\endgroup
\begin{abstract}

End-to-end driving methods learn from data collected from different drivers. Even in the same scene, different drivers may exhibit distinct driving styles when accelerating, braking, and interacting. During training, these style differences can create ambiguity for the model, leading to limited stable control at inference time. In this paper, we propose representing each style as a learnable style vector that defines a style direction for the planner. A vision-language-action planner is designed to provide continuous, composable driving-style control. During training, the learned style vectors guide a flow-matching decoder to generate feasible trajectories. At inference time, these style vectors can be scaled and combined for the flow-matching decoder without retraining. To quantify the resulting changes in driving behavior, we also introduce the Style Alignment Score (SAS) metric to measure human-referenced kinematic behavior after safety and task gating. On NAVSIM navtest, our PDMS remains above 0.90 across all six styles and their tested blends, reaching 0.954 for the best blend.

\end{abstract}

\section{Introduction}
\label{sec:intro}

\begin{figure}[t]
    \centering
    \includegraphics[width=0.94\linewidth]{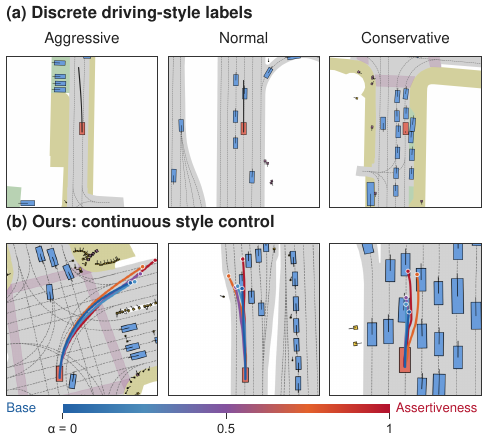}
    \caption{Driving-style control. (a) In near-stationary clips, discrete style labels~\cite{styledrive} may reflect scene conditions rather than driving style. (b) Within each scene, our method uses a style coefficient $\alpha$ to continuously adjust behavior from the base ($\alpha=0$, blue) toward a more assertive style ($\alpha=1$, red), without retraining.}
    \label{fig:teaser}
\end{figure}

Modern autonomous driving datasets are collected from many drivers, producing millions of training samples. Even in the same traffic situations, however, different drivers may show different driving styles when accelerating, braking, and interacting with other road users. Ground-truth trajectories with different driving styles can confuse end-to-end model training because the model does not explicitly distinguish between them ~\cite{transfuser,uniad,vad}. The learned behavior policy offers limited control over driving style, which leads to unstable trajectory predictions at inference time. In this paper, we propose making driving style explicitly adjustable within the same policy while preserving its ability to respond to traffic and complete the driving task.

StyleDrive~\cite{styledrive} proposed representing a driving policy with discrete style labels, such as aggressive, normal, and conservative. However, we find that the distributions of scenes and motion states can differ markedly across these style labels. 
For example, when style labels are assigned to low-speed or near-stationary clips, they reveal more about scene conditions
than about driving style (\cref{fig:teaser}). This makes it hard to distinguish style-relevant behavior from scene-relevant behavior. In this paper, we seek a behavioral reference that can be examined under comparable driving conditions. Specifically, we treat each trained policy as a distinct driver and characterize its style by the trajectories it generates across the same set of traffic scenes.

In this paper, our method transfers teacher-policy behaviors to a shared vision--language--action (VLA) planner and controls how they are expressed during inference. For each teacher, we represent its driving behavior as a task vector using parameter adaptation relative to a common base planner~\cite{taskvector}. For a given scene, the difference between the adapted and base representations defines the style direction. We use a heuristic coefficient to control the strength of this direction, allowing interpolation for each driving style separately. Directions from multiple teachers are combined through weighted addition, enabling extrapolation in the same representation space without retraining the shared planner. The combined driving styles guide a flow-matching decoder to generate control sequences as motion outputs, which are then integrated through a constrained motion model. Multiple kinematically feasible trajectory candidates are computed guided by the driving style. In addition, to quantify the resulting behavior, we introduce a Style Alignment Score (SAS), which measures interpretable, human-referenced kinematic features after safety and task gating. SAS allows style changes to be examined separately from planning performance, alongside the unchanged PDMS. 

Our NAVSIM evaluation covers both single-teacher control and multi-teacher composition, demonstrating composable performance with driving styles while maintaining high planning performance. Our contributions are summarized as follows:
\begin{itemize}
    \setlength{\topsep}{2pt}\setlength{\itemsep}{1pt}\setlength{\parsep}{0pt}
    
    \item We propose controlling driving style in the end-to-end method and formulating it as a learnable task vector conditioned on the data distribution.
    \item We design a VLA planner that enables driving-style control through a style-conditioned flow-matching decoder. The learned driving style can be scaled and combined at inference time without retraining to achieve composable performance.
    \item We propose the Style Alignment Score (SAS), which quantifies human-referenced kinematic behavior after safety and task gating, enabling comparison of driving style independently of planning performance.
    \item Our NAVSIM results indicate compositional capability and demonstrate its advantage using six teacher styles. PDMS remains above 0.90 for each driving style and reaches 0.954 with the best combination configurations.
\end{itemize}

\section{Related Work}
\label{sec:related}

\paragraph{End-to-end and VLA planning.}
End-to-end driving methods combine scene understanding and trajectory planning in a learned policy. TransFuser fuses camera and LiDAR features for imitation-based driving~\cite{transfuser}. UniAD organizes perception and prediction around planning, while VAD uses vectorized scene representations~\cite{uniad,vad}. VLA approaches use pretrained vision-language models to link visual observations and language instructions to action generation. EMMA represents driving inputs and outputs in a unified language format~\cite{emma}, whereas AutoVLA encodes trajectories as discrete feasible actions and predicts them autoregressively~\cite{autovla}. For multimodal trajectory generation, DiffusionDrive uses truncated diffusion with multimodal anchors~\cite{diffusiondrive}, while GoalFlow employs goal-conditioned flow matching~\cite{goalflow}. We build on a VLA planner and replace discrete action decoding with a continuous flow-matching head in a kinematically constrained control space. We focus on steering the resulting trajectory distribution with teacher-derived behavior directions.

\paragraph {Stylized autonomous driving.}
Stylized driving has been studied by learning driving styles from demonstrations and by modeling social preferences in vehicle interactions~\cite{kuderer,schwarting}. In end-to-end driving, StyleDrive introduces scenario-aware aggressive, normal, and conservative annotations, along with corresponding style-conditioned baselines~\cite{styledrive}. Its style-modulated PDMS adjusts comfort, progress, and time-to-collision terms to assess style alignment. Although these labels support categorical control, uneven scene and motion-state distributions across categories make individual preferences harder to interpret, as examined in our analysis (\cref{sec:exp:anc}). StyleVLA conditions trajectory generation on driving-style instructions with physics-informed training objectives~\cite{stylevla}. TrajHF fine-tunes generative trajectory models using human feedback to align their outputs with driving preferences~\cite{trajhf}. Our method focuses on deriving reusable behavior directions through teacher-policy distillation. Each teacher provides a behavioral reference across scenes, and its induced direction can be scaled and combined with other teacher directions within a shared VLA planner. We also retain PDMS and report human-referenced Kinematic behavior through SAS after safety and task gating, so behavioral changes can be examined separately from planning performance.

\paragraph{Task vectors and activation steering.}
Task arithmetic constructs task vectors by subtracting pretrained weights from their fine-tuned counterparts, allowing learned adaptations to be scaled and combined~\cite{taskvector}. Model Soups averages fine-tuned model weights~\cite{modelsoups}, while TIES-Merging addresses interference between parameter updates during merging~\cite{tiesmerging}. Activation steering instead intervenes in hidden representations during inference. ITI shifts selected attention-head activations~\cite{iti}, while style vectors identify activation directions associated with behavioral traits. Our method links parameter updates to representation-level control through teacher-policy distillation. The learned task vector induces a scene-dependent direction in the VLA planning representation. We combine teacher-specific task vectors and scale the resulting representation shift to condition trajectory generation.

\section{Method}
\label{sec:method}
\begin{figure*}[t]
    \centering
    \includegraphics[width=\linewidth]{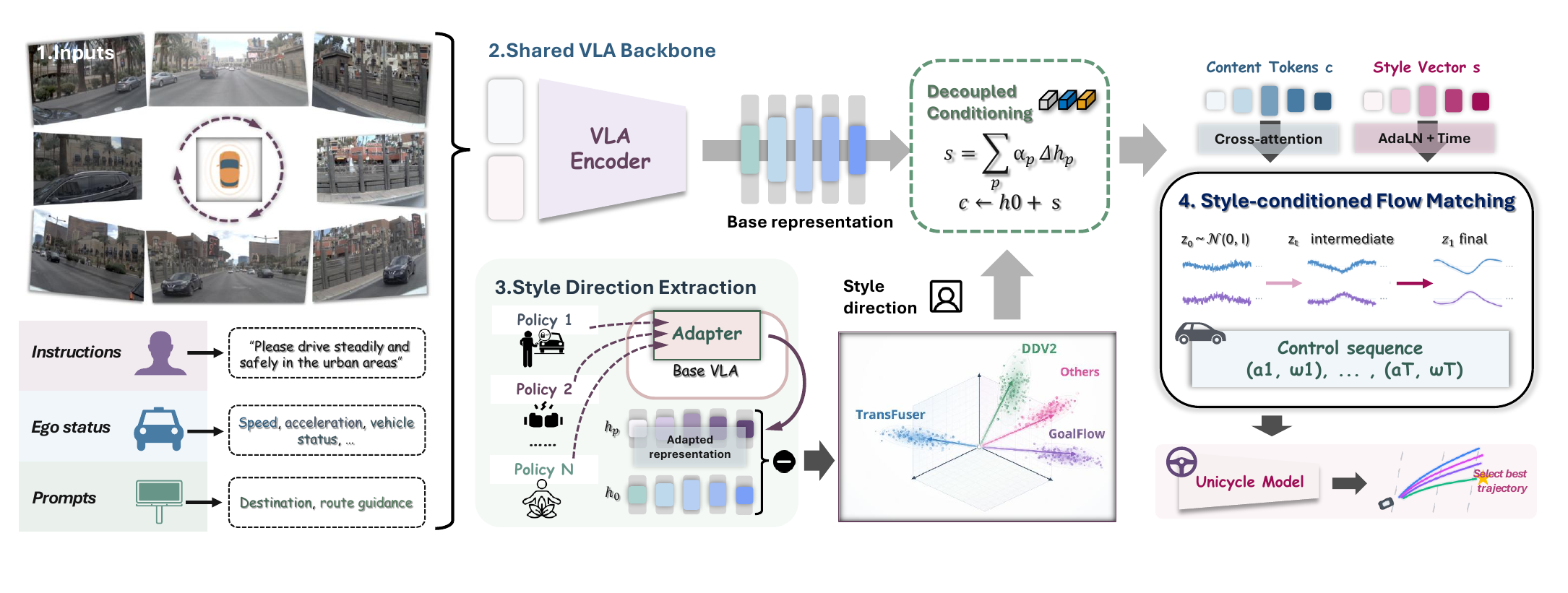}
    \caption{Overview of our pipeline. The VLA backbone provides content tokens. Styles are distilled through task-vector interpolation applied to those tokens, and a flow-matching decoder samples several feasible candidates before returning one.}
    \label{fig:pipeline}
\end{figure*}
\subsection{Overview}
\label{sec:method:overview}
The overall structure of our method is shown in \cref{fig:pipeline}. At each planning step, the planner takes multi-view images $I$, ego state $e$, language instructions, and the navigation prompt command $g$ as input. We use a pretrained VLA backbone to encode these inputs into content tokens. At the same time, task vectors are distilled from pretrained policies to represent the unique driving style through the extracted style direction. We then scale and combine the selected directions to condition a flow-matching decoder. The decoder generates trajectory candidates through a constrained motion model. The selected trajectory is represented as $\tau=\big((x_k,y_k,\psi_k)\big)_{k=1}^{T}$, where each step includes a location and heading. We use $T{=}10$ poses at intervals of $\Delta t{=}0.5$\,s.

\subsection{VLA backbone}
\label{sec:method:vla}
We use Qwen2.5-VL-3B~\cite{qwenvl} as the multimodal scene encoder to capture scene and command content. Multi-view camera images $I$ are preprocessed as visual tokens, while the ego state $e$ and navigation command $g$ are encoded as text tokens. A DIT backbone then jointly maps $x=(I,e,g)$ to a fixed set of content tokens $h_0(x)=f_{\theta_0}(x)\in\mathbb{R}^{L\times d}$. Here, $f_{\theta_0}$ denotes the pretrained VL encoder, whose frozen parameter $\theta_0$ covers only the encoder structure. The extracted content tokens encode both scene information and the navigation command for trajectory generation.

\subsection{Style as a direction in VLA content space}
\label{sec:method:style}
Based on the scene and command content, we propose adding a driving-style condition to the content token to enable style-controllable trajectory decoder performance. We employ several predefined policies, including GTRS~\cite{gtrs}, Hydra-MDP~\cite{hydramdp}, WoTe~\cite{wote}, TransFuser~\cite{transfuser}, GoalFlow~\cite{goalflow}, DiffusionDriveV2~\cite{diffusiondrivev2}, to build our teacher pool for distinguishing driving styles. For each teacher policy network $p$, we add a LoRA adapter~\cite{lora} to its learned content tokens while keeping the backbone and trajectory decoder frozen. The LoRA adaptor aligns the content token between the teacher networks and our basic VLA backbone. Following task arithmetic~\cite{taskvector}, we represent the task vector by the parameter update $\Delta W_p$ of the LoRA adaptor relative to the backbone parameter $\theta_0$. Then, for a scene input $x$, we can compute the base and teacher-adapted content token as $h_0(x)$ and $h_p(x)=f_{\theta_0+\Delta W_p}(x)$. The style direction $\Delta h_p(x)$ is conditioned on the difference for each teacher policy when handling different scenes.
\begin{equation}
    \Delta h_p(x) = h_p(x) - h_0(x).
    \label{eq:style-dir}
\end{equation}

\noindent\textbf{Scaling and composition.}
With a single driving teacher, the style direction can be scaled on the base content token as the style-conditioned input $c_{p,\alpha}(x)$ for the trajectory decoder
\begin{equation}
    c_{p,\alpha}(x) = h_0(x) + \alpha\,\Delta h_p(x),
    \qquad \alpha \in [0,1],
    \label{eq:interp}
\end{equation}
where $\alpha$ controls the influence of the extracted driving style. $\alpha{=}0$ indicates using only the base content, and $\alpha{=}1$ denotes fully adopting the style condition. Furthermore, with multiple teachers, we use a blending operator to combine the different task vectors with weights $\beta_p$. The blending process can be indicated as
\begin{equation}
    \begin{aligned}
    \Delta W_{\boldsymbol\beta}&=\sum_p\beta_p\Delta W_p,\\
    c_{\alpha,\boldsymbol\beta}(x)&=h_0(x)+
    \alpha\bigl(f_{\theta_0+\Delta W_{\boldsymbol\beta}}(x)-h_0(x)\bigr),
    \end{aligned}
    \label{eq:blend}
\end{equation}
where $f_{\theta_0+\Delta_W}$ represents the combined driving directions. We normalize the $\beta_p$ values so that they sum to 1, allowing multiple teachers to function as a single strong expert. Thus, the overall strength $\alpha$ remains independently adjustable as a single driver scale. The base VLA content token preserves the scene and navigation content for the decoder. Meanwhile, our composition method enriches the base content conditioned on the extracted driving style. The style-conditioned content tokens therefore enable a style update for the same trajectory generator without training a separate decoder or limiting the style to a fixed label. We conduct a careful ablation to determine the working scheme for scale and blend. The test results can be found in \cref{sec:exp:vector}.

\subsection{Flow-matching trajectory decoder}
\label{sec:method:fm}
With the style-conditioned content tokens, we use a flow-matching decoder to model multiple plausible trajectories~\cite{diffusiondrive,goalflow}. The decoder retains the basic multimodal generation capability while enabling style-controllable trajectory generation from continuous, composable style conditions. We use a DiT/AdaLN-based head~\cite{dit} with cross-attention to process the style-conditioned content tokens $c$ computed from \cref{eq:interp,eq:blend}. Under our scale and composition configuration, the original input to AdaLN can be kept, while the style direction is applied through the content tokens.

The decoder is learned in a standardized two-control-channel space. We predict $u_k=(a_k,\omega_k)$ for longitudinal acceleration and yaw rate, where each candidate corresponds to a sequence of control signals $z=(u_1,\dots,u_T)\in\mathbb{R}^{T\times 2}$. The training target $z_1$ is computed by inverting the target trajectory through the unicycle model into the standardized two control channels. The ground truth uses human trajectories for base training and the teacher trajectories for style distillation.

\noindent\textbf{Rectified-flow objective.}
Following rectified flow and flow matching design ~\cite{rectflow,flowmatching}, we construct
\begin{equation}
    z_t=(1-t)\,z_0+t\,z_1,\qquad t\in[0,1],
    \label{eq:interpolant}
\end{equation}
for the source $z_0\sim\mathcal{N}(0,I)$ and target $z_1$ distributions, using the target velocity $z_1-z_0$. Our decoder $v_\phi$ regresses the velocity through cross-attention with the style-conditioned content token $c$. The flow-matching training objective can be written as
\begin{equation}
    \mathcal{L}_{\mathrm{FM}}
    =\mathbb{E}_{t,\,z_0,\,z_1}\big\|\,v_\phi(z_t,t;c)-(z_1-z_0)\,\big\|^2 .
    \label{eq:fmloss}
\end{equation}
We also use a lightly weighted control-jerk penalty ($\lambda_j{=}0.01$) on the estimated clean controls, weighted by $t$ when flow times $t{>}0.5$. This objective is used for base training and style distillation. 

\noindent\textbf{Style Control for the Inference}
At inference, we use $N{=}16$ independent initial samples and compute $dot z_t=v_phi(z_t,t;c)$ with $K{=}100$ Euler steps. The update process can be written as
\begin{equation}
    z_{t+\frac1K}=z_t+\tfrac1K\,\dot z_t.
    \label{eq:odestep}
\end{equation}
With the sampled sets, we can represent different styles within the same scene, thereby obtaining a specific style condition rather than averaging them into a single point estimate.

To compute the final trajectory, we reverse the control normalization in the control space. Specifically, we clip the acceleration and yaw rate as $\bar a_k=\operatorname{clip}(a_k,-a_{\max},a_{\max})$ and $\bar\omega_k=\operatorname{clip}(\omega_k,-\omega_{\max},\omega_{\max})$. We use $a_{\max}{=}5$\,m/s$^2$, $\omega_{\max}{=}1$\,rad/s, and $v_{\max}{=}25$\,m/s. Given the initial ego speed $v_0$, heading $\psi_0=0$, and position $p_0=(0,0)$, the unicycle model processes the control signals as
\begin{equation}
\begin{aligned}
v_k
&= \operatorname{clip}\!\left(
v_0+\Delta t\sum_{i=1}^{k}\bar a_i,\,
0,v_{\max}\right),\\
\psi_k
&= \psi_{k-1}+\bar\omega_k\Delta t,\\
p_k
&= p_{k-1}
+v_k(\cos\psi_k,\sin\psi_k)\Delta t.
\end{aligned}
\label{eq:unicycle}
\end{equation}
The converted trajectory can be represented by $(v_k, \psi_k, p_k)$, producing candidates within the specified kinematic bounds. To select the best candidate, we use a prediction-based PDM selector~\cite{pdmclosed,navsim} that ranks the $N$ candidates using the current scene, map content, and constant-velocity forecasts of surrounding agents. The selected candidate is the final planner output and is evaluated on both style and planning performance.

\section{Experiments}
\label{sec:exp}

\begin{table*}[h]
    \centering
    \footnotesize
    \setlength{\tabcolsep}{6pt}
    \caption{We report PDMS and its sub-metrics: no-collision (NC), drivable-area compliance (DAC), ego-progress (EP), time-to-collision (TTC), comfort (Comf), and SAS assertiveness $H$ (\cref{eq:sas-axes}) on NAVSIM navtest. Each native teacher planner ($\alpha{=}T$) is paired with its distilled style ($\hookrightarrow$ ours) at full strength ($\alpha{=}1$). The last two rows show the highest PDMS for single-style and blended configurations among the evaluated settings. DDV2 denotes DiffusionDriveV2 as the base teacher. The base uses $\alpha{=}0$, and the human reference has $H{=}0$. All PDMS values are taken from the original papers \cite{transfuser,hydramdp,goalflow,diffusiondrivev2,wote}, except GTRS, whose v1 result we evaluated ourselves.}
    \label{tab:v1}
    \begin{tabular}{llccccccc}
        \toprule
        & & & \multicolumn{5}{c}{PDMS sub-scores} & SAS\\
        \cmidrule(lr){4-8}
        config & $\alpha$ & PDMS$\uparrow$ & NC & DAC & EP & TTC & Comf & $H$ \\
        \midrule
        human anchor & -- & 0.948 & 100.0 & 100.0 & 87.5 & 100.0 & 99.9 & 0.00 \\
        base (ours)  & 0   & 0.912 & 98.8 & 96.6 & 86.7 & 96.3 & \textbf{100.0} & 0.47 \\
        \midrule
        GTRS               & T   & 0.899 & 99.3 & 97.6 & 83.2 & 96.1 & \textbf{100.0} & $-0.01$ \\
        \quad$\hookrightarrow$ ours & 1.0 & 0.903 & 97.5 & 96.6 & 87.6 & 93.6 & \textbf{100.0} & 0.05 \\
        Hydra-MDP          & T   & 0.865 & 98.3 & 96.0 & 78.7 & 94.6 & \textbf{100.0} & $+0.05$ \\
        \quad$\hookrightarrow$ ours & 1.0 & 0.907 & 98.0 & 96.6 & 87.7 & 94.4 & \textbf{100.0} & 0.13 \\
        WoTe               & T   & 0.883 & 98.5 & 96.8 & 81.9 & 94.9 & 99.9  & 0.16 \\
        \quad$\hookrightarrow$ ours & 1.0 & 0.905 & 98.4 & 95.6 & 86.9 & 95.5 & \textbf{100.0} & 0.17 \\
        TransFuser         & T   & 0.840 & 97.7 & 92.8 & 79.2 & 92.8 & \textbf{100.0} & 0.06 \\
        \quad$\hookrightarrow$ ours & 1.0 & 0.900 & 98.6 & 95.7 & 85.8 & 95.5 & \textbf{100.0} & 0.11 \\
        GoalFlow           & T   & 0.903 & 98.4 & 98.3 & 85.0 & 94.6 & \textbf{100.0} & 1.91 \\
        \quad$\hookrightarrow$ ours & 1.0 & 0.935 & 99.4 & 97.9 & 89.0 & 98.0 & \textbf{100.0} & 1.97 \\
        DiffusionDriveV2   & T   & 0.912 & 98.3 & 97.9 & 87.5 & 94.8 & 99.9 & 2.70 \\
        \quad$\hookrightarrow$ ours & 1.0 & 0.941 & 99.5 & \textbf{98.5} & 89.5 & 98.1 & \textbf{100.0} & 3.24 \\
        \midrule
        DDV2 (ours) & 0.75 & 0.947 & 99.4 & 98.2 & 90.5 & 97.8 & \textbf{100.0} & 2.78 \\
        DDV2\,$+$\,GTRS (60/40) & 1.0 & \textbf{0.954} & \textbf{99.6} & \textbf{98.5} & \textbf{91.8} & \textbf{98.3} & \textbf{100.0} & 2.66 \\
        \bottomrule
    \end{tabular}
\end{table*}

\subsection{Setup}
\label{sec:exp:setup}
Our planner uses Qwen2.5-VL-3B~\cite{qwenvl} as the scene encoder. Multi-view images, ego status, and the navigation command are tokenized into content tokens that summarize both the scene and the requested maneuver. On top of that, we run the flow-matching decoder of \cref{sec:method:fm}, conditioned on the content tokens through cross-attention. It operates in a control space of longitudinal acceleration and yaw rate using a unicycle model, integrated with $K$ steps and sampled $N{=}16$ times per frame. We warm-start from the released \texttt{AutoVLA\_PDMS\_89} checkpoint~\cite{autovla} and jointly fine-tune the language backbone and decoder on NAVSIM navtrain. During fine-tuning, we keep the vision encoder frozen and use the human trajectory as the target. For the base result, we set $\alpha{=}0$.
Style adapter are distilled from teacher planners including GTRS~\cite{gtrs}, Hydra-MDP~\cite{hydramdp}, WoTe~\cite{wote}, TransFuser~\cite{transfuser}, GoalFlow~\cite{goalflow}, DiffusionDriveV2~\cite{diffusiondrivev2}, and applied through \cref{eq:interp}. Each adapter is realized by a LoRA~\cite{lora} module with $r{=}64$, $\alpha_{\mathrm{LoRA}}{=}128$, dropout $0.05$ for the attention and MLP units. As we freeze the base VLM, including the vision encoder and the flow head, only the LoRA weights are trained for each style adaptor. We evaluate our method on NAVSIM~\cite{navsim} navtest with about 12k scenarios, 8 shards, and report scenario-weighted PDMS (v1). We also present the Style Alignment Score (SAS), in which the policy is evaluated along an assertiveness axis $H$ and a proximity axis $P$ relative to the human (\cref{sec:exp:sas}).

\subsection{Style Alignment Score(SAS)}
\label{sec:exp:sas}
Safety and driving style answer different questions, so they should not be measured with the same metric. In this paper, we evaluate a policy with a two-step metric that uses gating and scoring. First, the gating step keeps only trajectories that satisfy NAVSIM's safety and task terms, including no collisions, drivable-area compliance, time-to-collision, and ego progress. We keep only trajectories whose style does not change the safety verdict. Second, we measure the alignment between the passing motion and style with a target driver. SAS and PDMS are therefore reported under different capacities. We leave PDMS unchanged and report SAS on it. The joint metric not only indicates statistical independence between the scores, but also shows whether a style that improves ego-progress can raise PDMS.

We represent a trajectory using physically interpretable kinematic atoms. The dynamic atoms include peak acceleration, deceleration, longitudinal and lateral jerk, $\{a^{+}, a^{-}, j_{\ell}, j_{t}\}$. The proximity atoms include min time-headway, time-to-collision, and object and VRU gap $\{\mathrm{thw},\mathrm{ttc},d_{\min},d_{\mathrm{vru}}\}$. For a policy $d$ and atom $f$, we take the per-scene value and its median $m_f(d)$ over the split. To compare performance against the human driver, we apply a signed $z$-score
\begin{equation}
  z_f(d) \;=\; \frac{m_f(d) - m_f(\mathrm{human})}{\sigma_f},
  \qquad
  \sigma_f = \operatorname{std}\nolimits_{d'} m_f(d'),
  \label{eq:sas-z}
\end{equation}
where $\sigma_f$ is computed over the evaluated policies, including the human, our base, and the six teachers. We define $z{=}0$ as human and $z{>}0$ as more than human. SAS measures the independent evidence of per-model separability and monotonic response to $\alpha$. The results show how far policies are on a fixed evaluation set. 

We characterize its kinematic expression using the harshness and assertiveness score $H$, which summarizes the intensity of acceleration, braking, and jerk. The proximity score $P$ quantifies how close the vehicle comes to other road users in space and time, based on distances, time headway, and time-to-collision. We aggregate the corresponding normalized metrics into these two scores:
\begin{equation}
  H(d)=\frac{1}{|\mathcal{D}|}\!\sum_{f\in\mathcal{D}} z_f(d),
  \qquad
  P(d)=\frac{1}{|\mathcal{P}|}\!\sum_{f\in\mathcal{P}} \big(-z_f(d)\big),
  \label{eq:sas-axes}
\end{equation}
where $\mathcal{D}$ denotes the dynamics atoms and $\mathcal{P}$ the proximity atoms. $H{>}0$ indicates harder, jerkier driving than the human, while $P{>}0$ indicates tighter gaps. SAS reports style as $(H,P)$, with $H$ as the primary validated axis. We show that $H$ is strongly separable by the model \cref{sec:exp:systematic} and is driven monotonically by the single strength $\alpha$ (\cref{sec:exp:style}). We also report a per-atom variance ratio $\eta^2$ alongside the axes, so the metric speaks for itself. The SAS metric has been shown to separate drivers while responding monotonically to a style control.

\subsection{Style and safety on navtest}
\label{sec:exp:v1}

We compare our method with the base policy and the pretrained six teachers in \Cref{tab:v1}, which reports PDMS and its sub-metrics on navtest, together with the SAS assertiveness axis $H$. The single-teacher result is computed at full strength ($\alpha{=}1$). The last two rows compare the highest PDMS values from scaling a single driving direction and blending multiple teachers across the evaluated settings. With single-teacher scaling, the style leads to variable performance. GTRS and Hydra-MDP show a smooth style close to human behavior with $H{\approx}0.05$--$0.15$, while DiffusionDriveV2 shows an aggressive style with $H{\approx}3.2$. However, their PDMS stays in a narrow band ($0.90$--$0.94$) around the base. The sub-metrics pinpoint the detailed changes. We can see that the hard-safety components, no-collision (NC) and drivable-area compliance (DAC), remain essentially flat, while ego-progress (EP) and time-to-collision (TTC) vary with $H$. The variable $H$ indicates that each native teacher has its own performance profile, while the scaled style guide provides a controllable range around it. Because each scaled style closely matches its native teacher in $H$, the adapter can transfer the specific style rather than a generic average. The detailed ablation of continuous strength control and style composition is examined in \cref{sec:exp:style}.

\subsection{Style is a Per-model Signal}
\label{sec:exp:systematic}
\begin{figure}[tbp]
    \centering
    \includegraphics[width=\linewidth]{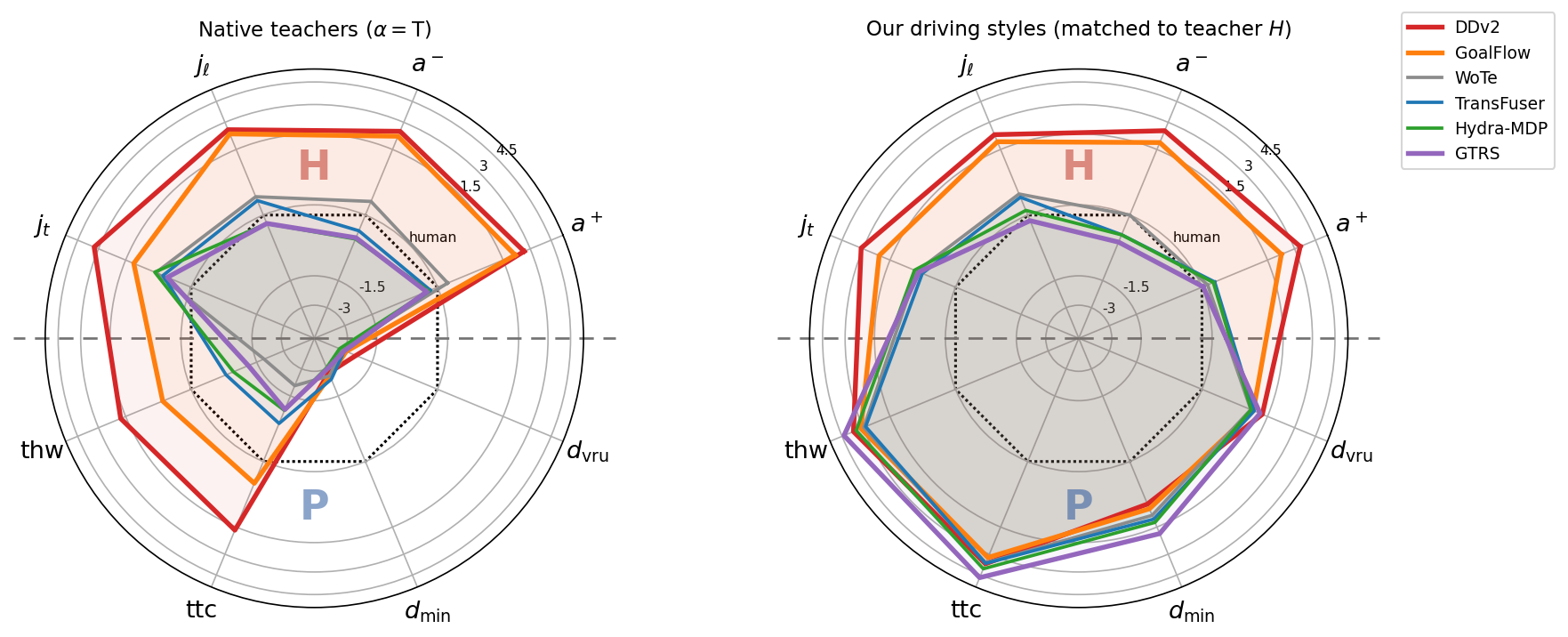}
    \caption{Comparison of style geometry in atom space. The dotted ring marks the human reference, with radius increasing as assertiveness rises on the dynamics half $H$ and gap-tightening on the proximity half $P$. Left: Native teachers. Right: Our scaled results.
    }
    \label{fig:radar}
\end{figure}

In this section, we show that our style is a real, systematic signal rather than scene noise. Using a self-consistent policy from pretrained end-to-end methods as expert drivers, we find that their styles are masked by kinematic signatures across scenes. As shown in \cref{tab:vardecomp}, we collect trajectories for 16 drivers, including the human reference, external planners, and our base model. We then perform a one-way variance decomposition of per-token kinematics, using driver style as the factor. The drivers have a mean $\eta^2$ of 0.27, reaching 0.46 for lateral jerk, which suggests that how a driver accelerates, brakes, and jerks is strongly driver-specific. The one exception with weak style is average speed ($\eta^2{=}0.005$), which is determined by road and traffic. These results show that our style is a continuous signal organized by model-as-driver rather than coarse scene labels. Furthermore, \cref{fig:radar} shows variation across policies in the dynamics atoms, demonstrating the assertiveness $H$ in the style-control analysis.

\begin{table}[tbp]
    \centering
    \small
    \setlength{\tabcolsep}{6pt}
    \caption{Variance decomposition of per-token kinematics with driver identity as the factor. Control features are driver-dominated, while average speed is scene-locked. In contrast, the coarse ANC label of~\cite{styledrive} carries little style signal once the environment is accounted for (\cref{sec:exp:anc}).}
    \label{tab:vardecomp}
    \begin{tabular}{lccc}
        \toprule
        kinematic feature & $\eta^2$(driver)$\uparrow$ & ICC(1) & $F$ \\
        \midrule
        avg.\ speed        & 0.005 & 0.005 & 20 \\
        peak accel         & 0.275 & 0.288 & 1631 \\
        peak decel         & 0.287 & 0.300 & 1735 \\
        long.\ jerk        & 0.445 & 0.461 & 3451 \\
        lat.\ jerk         & 0.460 & 0.476 & 3663 \\
        lat.\ accel (max)  & 0.167 & 0.176 & 862 \\
        \midrule
        mean               & 0.273 & 0.284 & -- \\
        \bottomrule
    \end{tabular}
\end{table}

\begin{figure*}[t]
    \vspace{-5pt}
    \centering
    \includegraphics[width=\linewidth]{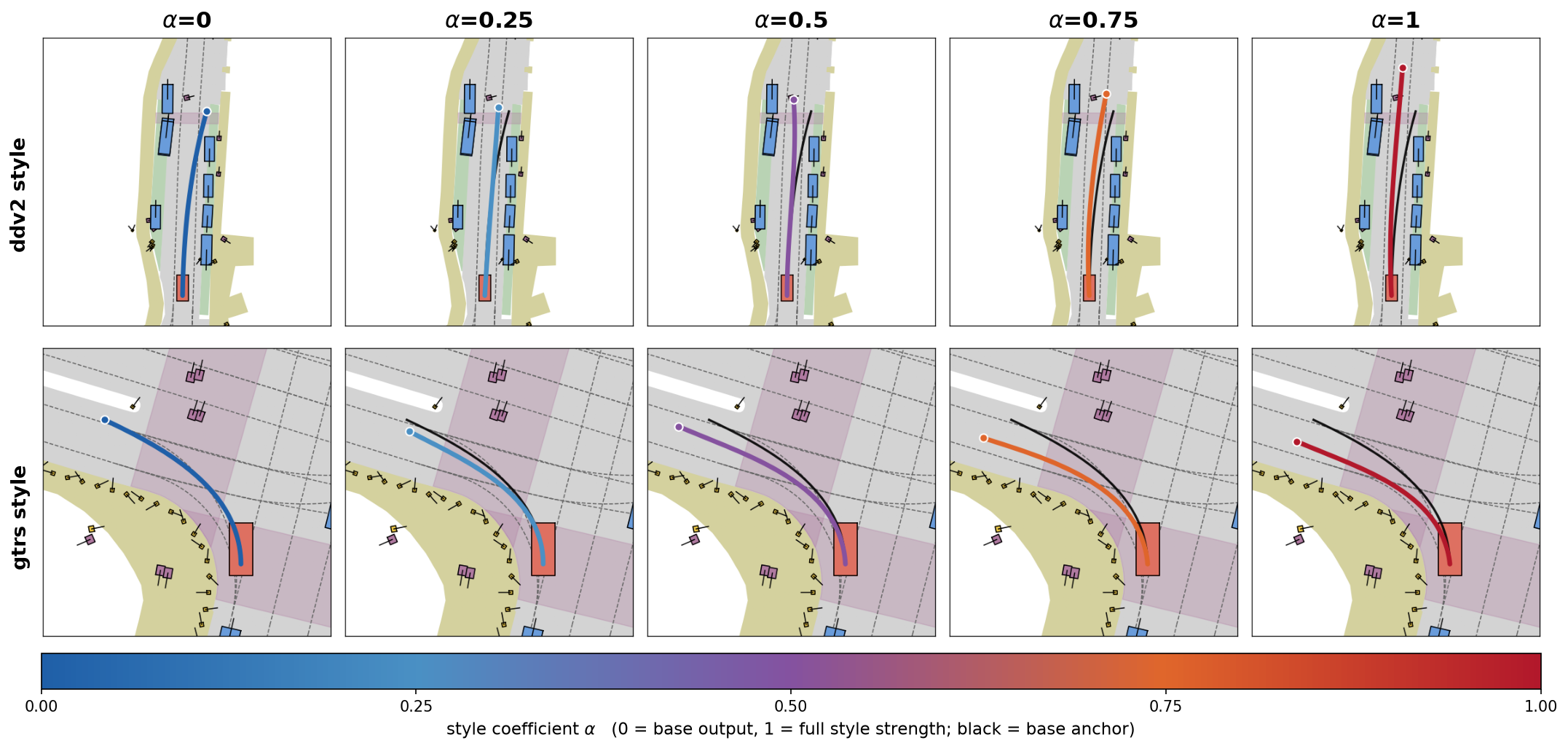}
    \caption{Style controllability in BEV. Each row shows one scene decoded at increasing style strength $\alpha=0\rightarrow1$. The top DiffusionDriveV gives an assertive style, while the bottom GTRS gives a cautious style. The black curve is the base result. Ego is outlined in red, and other agents in blue.}
    \label{fig:qual_sweep}
\vspace{-5pt}
\end{figure*}

\subsubsection{Limitation of Discrete Label}
\label{sec:exp:anc}
We assume that model-as-driver is the right unit for style, so the hand-assigned categorical labels should be evaluated with the same kinematic atoms. We use the aggressive/normal/conservative (ANC) labels released by StyleDrive~\cite{styledrive} for the navtest scenes and evaluate them with our kinematic atoms on each human-collected trajectory. StyleDrive assigns ANC based on the ego kinematics, so measuring the labels against those same kinematics asks exactly what the labels are intended to encode. Grouping the human trajectories by ANC, the raw between-group effect size on the assertiveness atoms is near zero ($\varepsilon^2{=}0.006$ for peak acceleration, $0.012$ for peak deceleration, $\le 0.05$ for jerk, $0.063$ aggregated to $H$). Only the average speed $\varepsilon^2{=}0.088$ differs, and the proximity atoms seem to affect speed directly. Grouping the same trajectories by speed recovers large effects on exactly these atoms ($\varepsilon^2{=}0.38$--$0.40$ for peak accel/decel), which indicates that the near-zero effect of ANC reflects the true performance. When we regress each atom on the scene environment (scenario, turn, lead, road), ANC adds almost nothing ($\Delta R^2(\text{ANC}\mid\text{env}){=}0.037$ on $H$ and $0.000$ on $P$. \cref{tab:anc}). In contrast, the same regression shows our method has a clear difference $\Delta R^2{=}0.232$ on $H$, over six times that of the label.

We find that the ANC is dominated by speed. Its composition follows a six-bin speed partition (Cramér's$V{=}0.34$), and the conservative is largely a stationary-scene detector, as shown by the median speed$0.45$\,m/s, $77.5\%$ of its scenes being near-stationary ($<2$\,m/s), and $87\%$ of fully stopped scenes. A plain speed partition with no style design separates our atoms better than ANC on nearly every atom. The label thus re-describes driving content by speed and scene, rather than using an independent per-model style.

\begin{table}[tbp]
    \centering
    \small
    \setlength{\tabcolsep}{5pt}
    \caption{Two complementary diagnostics of driving styles on the navtest. The ANC column groups the human log-replay trajectories by ANC label. The driver column groups the 16 policies by identity. Top: Raw between-group effect size. Bottom: Incremental adjustment of $R^2$ after regressing out the scene environment, penalizing the differing degrees of freedom.}
    \label{tab:anc}
    \begin{tabular}{lcc}
        \toprule
        & ANC label & driver identity \\
        \midrule
        raw effect size, peak accel        & 0.006 & 0.275 \\
        raw effect size, peak decel        & 0.012 & 0.287 \\
        raw effect size, assertiveness $H$ & 0.063 & 0.270 \\
        \midrule
        $\Delta R^2(\cdot\mid\text{env})$ on $H$ & 0.037 & 0.232 \\
        $\Delta R^2(\cdot\mid\text{env})$ on $P$ & 0.000 & 0.065 \\
        \bottomrule
    \end{tabular}
\end{table}

\subsection{Controllable and Composable Style}
\label{sec:exp:style}

\begin{figure}[tbp]
\vspace{-10pt}
    \centering
    \includegraphics[width=\linewidth]{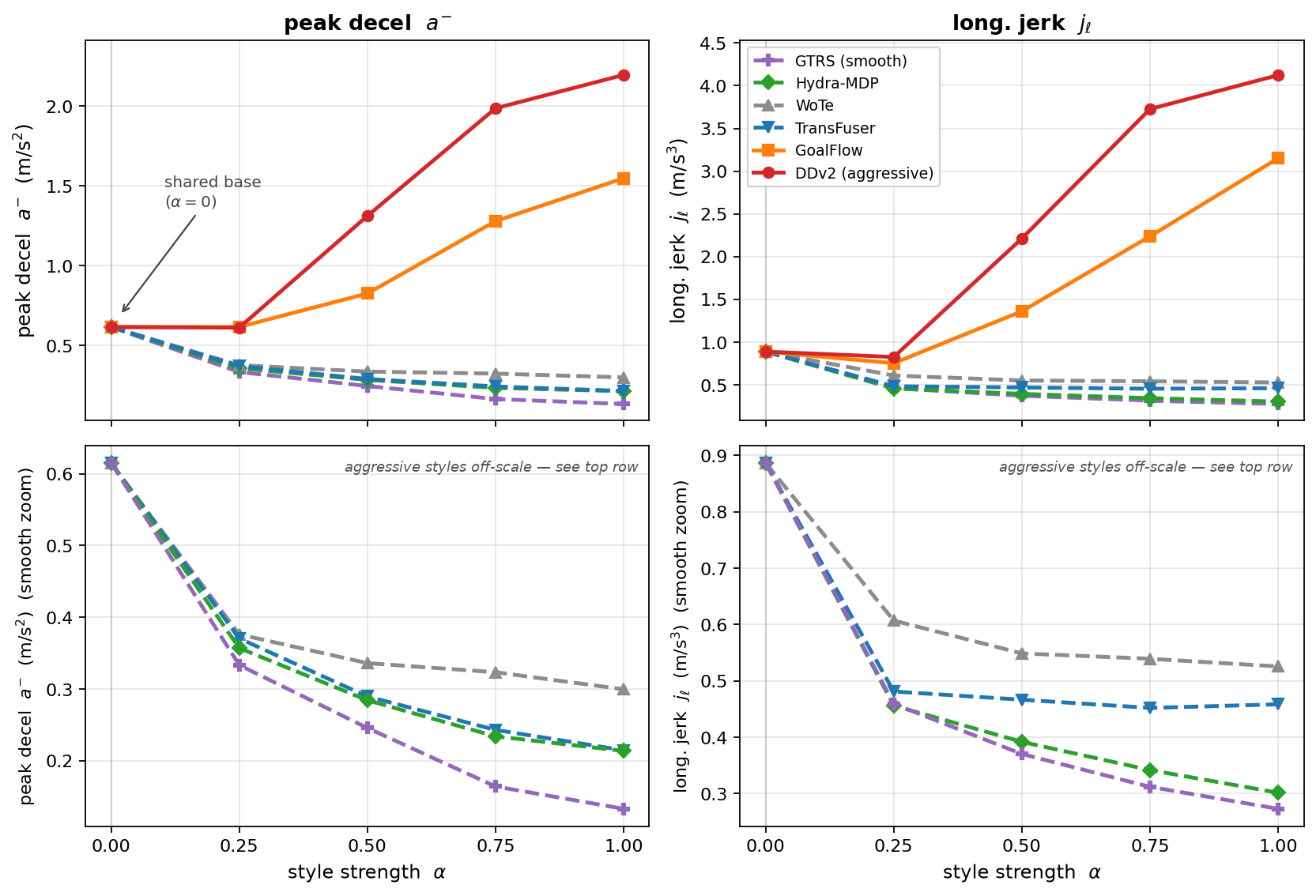}
    \caption{As the single strength $\alpha$ increases, the peak deceleration and longitudinal jerk move monotonically from the base toward each teacher operating point. $\alpha$ is a continuous style coefficient rather than a discrete selector. The smoother styles separate clearly at small magnitudes.}
    \label{fig:dose}
    
\end{figure}

As shown in \cref{tab:following,fig:dose}, sweeping $\alpha$ shifts the assertiveness axis $H$ and its kinematics from the base toward each teacher, so one coefficient spans a range instead of switching between presets. For the DiffusionDriveV2 style, increasing $\alpha$ from 0 to 1 raises $H$ monotonically to $3.24$ while PDMS stays high and peaks near $\alpha{=}0.75$, indicating that strong style expression is achieved at no cost to safety. As shown in \cref{fig:dose}, the dose-response remains smooth and monotonic across the full style set, demonstrating that multi-modality is common enough to matter on navtest. Among scenes with sufficient driver coverage, about 7.4\% are laterally bimodal and 13.9\% longitudinally bimodal. From the visualization of decoded trajectories in \cref{fig:qual_sweep}, we can see that the single-mode regressor returns a path between the modes. The sampling decoder draws distinct maneuvers and commits to one, so styles are kept even with averaging.

\begin{table}[tbp]
    \centering
    \small
    \setlength{\tabcolsep}{4pt}
    \caption{Style shifts as $\alpha$ increases. A single coefficient, $\alpha$, steers the base toward each teacher in opposite directions along the assertiveness axis $H$. Left: the aggressive DiffusionDriveV2 style. Right: the smooth GTRS style.}
    \label{tab:following}
    \begin{minipage}[t]{0.49\linewidth}
        \centering
        \begin{tabular}{lcc}
            \toprule
            $\alpha$ & PDMS$\uparrow$ & $H$ \\
            \midrule
            0.00 & 0.912 & 0.47 \\
            0.25 & 0.926 & 0.53 \\
            0.50 & 0.941 & 1.50 \\
            0.75 & \textbf{0.947} & 2.78 \\
            1.00 & 0.941 & 3.24 \\
            \bottomrule
        \end{tabular}
        \subcaption{DiffusionDriveV2 (aggressive): $H$ climbs to $3.24$.}
    \end{minipage}\hfill
    \begin{minipage}[t]{0.49\linewidth}
        \centering
        \begin{tabular}{lcc}
            \toprule
            $\alpha$ & PDMS$\uparrow$ & $H$ \\
            \midrule
            0.00 & 0.912 & 0.47 \\
            0.25 & 0.919 & 0.27 \\
            0.50 & \textbf{0.923} & 0.19 \\
            0.75 & 0.916 & 0.10 \\
            1.00 & 0.903 & 0.05 \\
            \bottomrule
        \end{tabular}
        \subcaption{GTRS (smooth): $H$ falls as driving turns more conservative.}
    \end{minipage}
\end{table}

\noindent\textbf{Properties of the Style Direction.}
\label{sec:exp:vector}%
The linear attribute in content space may not be consistent in the measured output space. So we test these two task-vector properties directly on the style axis. Linearity is demonstrated by the dose-response as in \cref{sec:exp:style}, where$H$ increases monotonically with $\alpha$. In addition, we quantify it as the fit $R^2$ of $H$ against $\alpha$ over the full sweep as shown in \cref{fig:dose,tab:following}. The composition ability indicates whether a blend $\sum_i w_i \Delta_i$ matches the weighted sum of the single-style shifts. \cref{tab:vector} shows our styles to reach new operating points beyond any single teacher.

We also show that a convex blend creates a new operating point at no extra training cost and reaches regions that no single teacher can. \Cref{tab:blends} lists measured blends at $\alpha{=}1$. The strongest DiffusionDriveV2/GTRS result achieves $\mathrm{PDMS}{=}0.954$, surpassing the highest-PDMS single-style setting across the evaluated strength sweep, $0.947$, as shown in \cref{tab:v1}. 
Besides, the blends without DiffusionDriveV2 have a narrower band of $H{\approx}1.05$, indicating a continuous range of styles with uniformly high safety. Same-direction blends follow the weighted-sum prediction within $\sim\!10\%$, which shows that aggressive$+$smooth blends are super-additive. Because the aggressive direction dominates, the measured $H$ exceeds the linear prediction instead of averaging toward the milder teacher.

\begin{table}[tbp]
    \centering
    \caption{Style directions behave as reusable, composable task vectors.}
    \label{tab:taskvec}

    \footnotesize
    \begin{subtable}{\columnwidth}
        \centering
        \subcaption{Task-vector properties. The $R^2$ of $H$ versus $\alpha$ shows a linear relationship across the aggressive teachers, where $H$ has room to move.}
        \label{tab:vector}

        \setlength{\tabcolsep}{4pt}
        \begin{tabular}{@{}llc@{}}
            \toprule
            property & measurement & result \\
            \midrule
            linearity in $\alpha$ & $R^2$ of $H$ vs.\ $\alpha$ & $0.93$ \\
            additive composition & error vs.\ weighted sum (same-dir.) & $6.07\%$ \\
            \bottomrule
        \end{tabular}
    \end{subtable}

    \vspace{4pt}
    \begin{subtable}{\columnwidth}
        \centering
        \subcaption{Representative blends at $\alpha{=}1$, shown as a convex mixture $\sum_i w_i\Delta_i$ of single-style directions. ddv2 = DiffusionDriveV2, gf = GoalFlow, tf = TransFuser, gtrs = GTRS, hy = Hydra-MDP.}
        \label{tab:blends}

        \setlength{\tabcolsep}{5pt}
        \begin{tabular}{@{}lccc@{}}
            \toprule
            style blend (weights) & PDMS$\uparrow$ & $H$ & $P$ \\
            \midrule
            ddv2\,$+$\,gtrs (60/40)
                & \textbf{0.954} & $+2.66$ & $+2.35$ \\
            ddv2\,$+$\,gf\,$+$\,hy (40/40/20)
                & 0.950 & $+2.40$ & $+2.27$ \\
            ddv2\,$+$\,gf (60/40)
                & 0.949 & $\mathbf{+2.94}$ & $+2.16$ \\
            ddv2\,$+$\,gf\,$+$\,tf\,$+$\,gtrs (35/35/15/15)
                & 0.940 & $+2.11$ & $+2.33$ \\
            gf\,$+$\,gtrs (50/50)
                & 0.930 & $+1.04$ & $+2.20$ \\
            tf\,$+$\,gf (50/50)
                & 0.921 & $+1.05$ & $+2.12$ \\
            \bottomrule
        \end{tabular}
    \end{subtable}
\end{table}

\subsection{Ablation Study}
\label{sec:exp:ablate}
In this section, we study how the trajectory decoder design and the inference hyperparameter settings affect controllable style. All training uses navtrain, and all variants are evaluated on the same navtest split. We compare decoder variants at matched capacity using PDMS and its safety sub-metrics. More reports about the reachable style span $\Delta H$ of the range of the assertiveness axis that a single model covers across its style are shown in \cref{sec:suppl:ablate} and the full sweep in \cref{sec:suppl:v1full}. For the decoder ablation, we replace the flow-matching decoder with the discrete action codebook, while keeping conditioning and training data fixed. The codebook preserves safety, but it quantizes trajectories to a fixed set of anchors, limiting the reachable style range and leaving only coarse, quantized style control as shown in \cref{tab:ablate_dec}-a. In contrast, the flow-matching decoder restores a smooth, broader style axis at the same safety level. As for the inference hyperparameter, we set the two inference knobs. As shown in  \cref{tab:ablate_sweep}, both PDMS and the style $\Delta H$ are saturate at $N{=}16$, $K{=}20$. We find that controllability lies in the style direction rather than in the sampler, so we default to $N{=}16$ and $K{=}100$. We also note our limitations. SAS is validated on assertiveness $H$, while proximity $P$ is reported only as a secondary, preliminary axis. On \textsc{navtest}, proximity atoms vary much less across policies than dynamics atoms, where cross-policy std is $0.18$-$0.31$ vs $0.57$-$1.55$. This result shows that the style easily collapses on this split, dominated by open-road, low-interaction scenes. Within this narrow range, every native teacher could loosen gaps by $P{<}0$ while every style tightens them $P{>}0$ as \cref{fig:radar}. So the style direction may not be able to reproduce the base performance on $P$. With the collapsed style, it remains open whether $P$ is independently controllable. Resolving this requires an interaction-dense, reactive closed-loop world model, which could be left to future work.

\begin{table}[tbp]
    \centering
    \footnotesize
    \caption{Ablations on decoder comparison (a) and inference budget (b).}
    \label{tab:ablate}
    \begin{subtable}{\columnwidth}
        \centering
        \setlength{\tabcolsep}{6pt}
        \subcaption{Decoder Comparison. Both hard safety (NC/DAC) and PDMS are close, but the discrete action limits the reachable style range to coarse, quantized control. Flow matching yields a smoother, broader style axis at equal safety.}
        \label{tab:ablate_dec}
        \resizebox{\columnwidth}{!}{%
        \begin{tabular}{@{}lccccccc@{}}
            \toprule
            decoder & PDMS$\uparrow$ & NC$\uparrow$ & DAC$\uparrow$ & EP$\uparrow$ & TTC$\uparrow$ & Comf$\uparrow$ & $\Delta H$ \\
            \midrule
            discrete codebook    & 0.891 & 0.984 & 0.956 & 0.818 & 0.980 & 0.999 & 0.8 \\
            flow matching (ours) & 0.912 & 0.988 & 0.966 & 0.867 & 0.963 & 1.000 & 3.2 \\
            \bottomrule
        \end{tabular}}

    \end{subtable}

    \vspace{4pt}
    \begin{subtable}{\columnwidth}
        \centering
        \setlength{\tabcolsep}{5pt}
        \subcaption{Inference budget. PDMS and $\Delta H$ saturate by $N{=}16$, $K{=}20$}
        \label{tab:ablate_sweep}
        \begin{tabular}{@{}lcc@{}}
            \toprule
            setting & PDMS$\uparrow$ & $\Delta H$ \\
            \midrule
            \multicolumn{3}{@{}l}{\emph{candidates $N$ (steps $K{=}100$)}}\\
            $N{=}4$          & 0.883 & 2.5 \\
            $N{=}8$          & 0.899 & 3.1 \\
            $N{=}16$ (ours)  & 0.912 & 3.2 \\
            $N{=}64$         & 0.913 & 3.2 \\
            \midrule
            \multicolumn{3}{@{}l}{\emph{steps $K$ (candidates $N{=}16$)}}\\
            $K{=}5$          & 0.864 & 2.3 \\
            $K{=}20$         & 0.910 & 3.2 \\
            $K{=}100$ (ours) & 0.912 & 3.2 \\
            \bottomrule
        \end{tabular}
    \end{subtable}
\end{table}

\section{Conclusion}
\label{sec:conclusion}


We proposed controlling the driving style in the end-to-end method and formulating it as a learnable task vector conditioned on the data distribution. Each learned task vector represents a scene-dependent style direction in the planner's content tokens. With the style task vectors, we presented a VLA planner that enables driving-style control via a style-conditioned flow-matching decoder. The learned driving style can be scaled and combined at inference without retraining to reach composable performance. We also introduced the Style Alignment Score (SAS), which quantifies human-referenced kinematic behavior after safety and task gating, enabling comparison of driving style independent of planning performance. Our NAVSIM results indicated the composable ability and demonstrated its priority using six teacher styles. PDMS remained above 0.90 with each driving style and reached 0.954 with the best combination configurations.
\section*{LLM Usage Statement}
No LLMs or AI tools were used to obtain reviewer-style feedback on the full manuscript or major parts of it.
We used ChatGPT (OpenAI) to assist with English
language polishing, literature organization,
and code debugging. All scientific ideas,
methodological designs, experimental results,
and final manuscript content were reviewed
and verified by the authors.

{
    \small
    \bibliographystyle{ieeenat_fullname}
    \bibliography{main}
}

\clearpage
\appendix

\section{Full Style Control and Composition Results}
\label{sec:suppl:v1full}
\cref{tab:v1full_single,tab:v1full_combo} report the full results for individual style sweeps and style blends, respectively. For style sweeps and blends, PDMS is the scenario-weighted mean across eight \textsc{navtest} shards. Native-teacher PDMS values are shown in \cref{tab:v1}. In addition to PDMS and the SAS axes $H$ and $P$, we include the individual kinematic and proximity measurements, including dimensions with weak separation on this split. The \texttt{scope} column distinguishes human and base references (\texttt{anchor}), individual style sweeps (\texttt{sweep}), and blends (\texttt{combo}). In \texttt{sweep}, $\alpha$ varies the strength of a single style, and $\alpha{=}\mathrm{T}$ denotes the native teacher. In \texttt{combo}, the configuration name gives the mixing weights of the teacher task vectors, while $\alpha$ controls their overall strength.

The full individual sweeps in \cref{tab:v1full_single} extend the selected examples in \cref{tab:following}. For all six styles, $H$ changes monotonically with $\alpha$, though in different directions. DiffusionDriveV2 raises $H$ from $0.47$ to $3.24$, while GTRS lowers it from $0.47$ to $0.05$. The corresponding peak acceleration and longitudinal jerk rise from $0.53$ and $0.89$ to $2.66$ and $4.12$ for DiffusionDriveV2, but fall to $0.24$ and $0.27$ for GTRS. Average speed varies less than these dynamics measurements, consistent with its weak driver-specific signal in \cref{sec:exp:systematic}. PDMS stays above $0.90$ across the tested style strengths, but does not change monotonically with $\alpha$. The proximity axis $P$ also does not follow the native teachers: it is positive for the distilled styles and negative for the native teachers. 

The blend results in \cref{tab:v1full_combo} provide a fuller view of the composition reported in \cref{tab:blends}. Changing the mixing weights can affect both $H$ and PDMS. For DiffusionDriveV2/TransFuser at $\alpha{=}1$, increasing the DiffusionDriveV2 weight from $30\%$ to $70\%$ changes $H$ from $0.73$ to $2.89$ and PDMS from $0.901$ to $0.948$. A fixed blend can also be scaled: DiffusionDriveV2/GTRS (60/40) moves from $H{=}0.47$ at $\alpha{=}0$ to $0.91$ at $\alpha{=}0.5$ and $2.66$ at $\alpha{=}1$, with PDMS reaching $0.954$. These output changes are not necessarily additive, consistent with the nonlinear blended representation in \cref{sec:method:style}. All tested blends retain PDMS above $0.90$.

\begin{table*}[t]
    \centering
    \scriptsize
    \setlength{\tabcolsep}{3pt}
    \resizebox{\textwidth}{!}{%
    \begin{tabular}{ll c c cc c cccc ccccc}
        \toprule
        scope & teacher & $\alpha$ & PDMS$\uparrow$ & $H$ & $P$ & $\bar v$ & $a^{+}$ & $a^{-}$ & $j_{\ell}$ & $j_{t}$ & thw & ttc & lead & $d_{\min}$ & $d_{\mathrm{vru}}$ \\
        \midrule
        anchor & \texttt{human} & -- & 0.9480 & +0.00 & +0.00 & 4.6181 & 0.2446 & 0.3059 & 0.2898 & 0.1026 & 3.7850 & 4.3915 & 12.9060 & 4.5340 & 12.0630 \\
        anchor & \texttt{base(ours)} & -- & 0.9118 & +0.47 & +0.07 & 4.6051 & 0.5282 & 0.6151 & 0.8865 & 0.4750 & 3.7640 & 4.3710 & 12.8275 & 4.5030 & 12.0780 \\
        sweep & \texttt{gtrs} & 0 & 0.9118 & +0.47 & +0.07 & 4.6051 & 0.5282 & 0.6151 & 0.8865 & 0.4750 & 3.7640 & 4.3710 & 12.8275 & 4.5030 & 12.0780 \\
        sweep & \texttt{gtrs} & 0.25 & 0.9187 & +0.27 & +1.69 & 4.8300 & 0.3713 & 0.3331 & 0.4590 & 0.5258 & 3.2140 & 3.5045 & 11.6580 & 4.4240 & 11.9630 \\
        sweep & \texttt{gtrs} & 0.5 & 0.9230 & +0.19 & +1.96 & 4.9096 & 0.3341 & 0.2454 & 0.3701 & 0.4722 & 3.1210 & 3.3810 & 11.5660 & 4.3940 & 11.9520 \\
        sweep & \texttt{gtrs} & 0.75 & 0.9158 & +0.10 & +2.28 & 4.9346 & 0.2753 & 0.1638 & 0.3118 & 0.4139 & 3.0620 & 3.2390 & 11.1695 & 4.3610 & 11.8570 \\
        sweep & \texttt{gtrs} & 1 & 0.9030 & +0.05 & +2.86 & 4.9977 & 0.2449 & 0.1326 & 0.2728 & 0.3721 & 2.8625 & 3.0440 & 10.8545 & 4.2760 & 11.7870 \\
        sweep & \texttt{gtrs} & T & 0.899 & -0.01 & -1.70 & 4.4350 & 0.2160 & 0.1840 & 0.2520 & 0.2160 & 4.0270 & 4.6615 & 14.1960 & 4.9570 & 12.7700 \\
        sweep & \texttt{hydra\_mdp} & 0 & 0.9118 & +0.47 & +0.07 & 4.6051 & 0.5282 & 0.6151 & 0.8865 & 0.4750 & 3.7640 & 4.3710 & 12.8275 & 4.5030 & 12.0780 \\
        sweep & \texttt{hydra\_mdp} & 0.25 & 0.9191 & +0.28 & +1.62 & 4.8250 & 0.3668 & 0.3568 & 0.4559 & 0.5304 & 3.2820 & 3.5350 & 11.7510 & 4.3980 & 11.9590 \\
        sweep & \texttt{hydra\_mdp} & 0.5 & 0.9232 & +0.21 & +1.72 & 4.8747 & 0.3171 & 0.2840 & 0.3916 & 0.5004 & 3.2250 & 3.4960 & 11.6775 & 4.4050 & 11.9580 \\
        sweep & \texttt{hydra\_mdp} & 0.75 & 0.9198 & +0.16 & +2.03 & 4.8833 & 0.2932 & 0.2337 & 0.3416 & 0.4722 & 3.1280 & 3.3625 & 11.3740 & 4.3740 & 11.9080 \\
        sweep & \texttt{hydra\_mdp} & 1 & 0.9067 & +0.13 & +2.30 & 4.9509 & 0.2756 & 0.2133 & 0.3013 & 0.4412 & 3.0530 & 3.2420 & 11.2290 & 4.3500 & 11.8650 \\
        sweep & \texttt{hydra\_mdp} & T & 0.8650 & +0.05 & -1.64 & 4.5170 & 0.2250 & 0.1700 & 0.2510 & 0.3690 & 3.9170 & 4.6515 & 14.1240 & 4.9690 & 12.8480 \\
        sweep & \texttt{wote} & 0 & 0.9118 & +0.47 & +0.07 & 4.6051 & 0.5282 & 0.6151 & 0.8865 & 0.4750 & 3.7640 & 4.3710 & 12.8275 & 4.5030 & 12.0780 \\
        sweep & \texttt{wote} & 0.25 & 0.9164 & +0.32 & +1.57 & 4.8201 & 0.4228 & 0.3759 & 0.6068 & 0.5135 & 3.2790 & 3.5840 & 11.6820 & 4.4490 & 11.8890 \\
        sweep & \texttt{wote} & 0.5 & 0.9192 & +0.29 & +1.66 & 4.8399 & 0.3552 & 0.3358 & 0.5485 & 0.5483 & 3.2160 & 3.4620 & 11.6325 & 4.4530 & 11.9950 \\
        sweep & \texttt{wote} & 0.75 & 0.9152 & +0.23 & +1.81 & 4.8764 & 0.3007 & 0.3231 & 0.5390 & 0.4679 & 3.2080 & 3.4450 & 11.4520 & 4.3930 & 11.9430 \\
        sweep & \texttt{wote} & 1 & 0.9051 & +0.17 & +2.00 & 4.8702 & 0.2523 & 0.2993 & 0.5254 & 0.4028 & 3.1610 & 3.3650 & 11.4460 & 4.3900 & 11.8680 \\
        sweep & \texttt{wote} & T & 0.8830 & +0.16 & -2.10 & 4.3269 & 0.2693 & 0.3496 & 0.4682 & 0.3516 & 4.1780 & 4.9650 & 14.3820 & 4.8985 & 12.8190 \\
        sweep & \texttt{ddv2} & 0 & 0.9118 & +0.47 & +0.07 & 4.6051 & 0.5282 & 0.6151 & 0.8865 & 0.4750 & 3.7640 & 4.3710 & 12.8275 & 4.5030 & 12.0780 \\
        sweep & \texttt{ddv2} & 0.25 & 0.9262 & +0.53 & +1.49 & 4.7682 & 0.5200 & 0.6112 & 0.8234 & 0.6351 & 3.2960 & 3.6320 & 11.8680 & 4.4540 & 11.9040 \\
        sweep & \texttt{ddv2} & 0.5 & 0.9406 & +1.50 & +1.45 & 4.8019 & 1.4282 & 1.3115 & 2.2109 & 1.0144 & 3.2270 & 3.6350 & 12.1850 & 4.4710 & 12.0180 \\
        sweep & \texttt{ddv2} & 0.75 & 0.9467 & +2.78 & +2.20 & 4.8995 & 2.5549 & 1.9862 & 3.7266 & 1.8909 & 3.0130 & 3.3460 & 11.8520 & 4.4440 & 11.7650 \\
        sweep & \texttt{ddv2} & 1 & 0.9407 & +3.24 & +2.23 & 4.8513 & 2.6606 & 2.1951 & 4.1235 & 2.5209 & 3.0275 & 3.2980 & 12.0275 & 4.4460 & 11.7610 \\
        sweep & \texttt{ddv2} & T & 0.9120 & +2.70 & -0.15 & 4.8577 & 2.0381 & 1.9625 & 4.1628 & 1.9892 & 3.4170 & 3.9205 & 13.5680 & 4.9230 & 12.6070 \\
        sweep & \texttt{goalflow} & 0 & 0.9118 & +0.47 & +0.07 & 4.6051 & 0.5282 & 0.6151 & 0.8865 & 0.4750 & 3.7640 & 4.3710 & 12.8275 & 4.5030 & 12.0780 \\
        sweep & \texttt{goalflow} & 0.25 & 0.9235 & +0.50 & +1.37 & 4.7899 & 0.4476 & 0.6154 & 0.7507 & 0.6462 & 3.3200 & 3.6700 & 11.8880 & 4.4930 & 11.9030 \\
        sweep & \texttt{goalflow} & 0.5 & 0.9191 & +0.85 & +1.56 & 4.7600 & 0.7489 & 0.8253 & 1.3588 & 0.8111 & 3.3030 & 3.6760 & 11.8095 & 4.4290 & 11.7950 \\
        sweep & \texttt{goalflow} & 0.75 & 0.9256 & +1.49 & +1.66 & 4.8250 & 1.3531 & 1.2791 & 2.2369 & 1.0610 & 3.2000 & 3.5730 & 11.7940 & 4.4070 & 11.9710 \\
        sweep & \texttt{goalflow} & 1 & 0.9354 & +1.97 & +1.93 & 4.8689 & 1.7530 & 1.5476 & 3.1498 & 1.2741 & 3.1210 & 3.4700 & 11.6940 & 4.4230 & 11.8470 \\
        sweep & \texttt{goalflow} & T & 0.9030 & +1.91 & -0.96 & 4.6061 & 1.6850 & 1.7739 & 3.7930 & 0.7634 & 3.7250 & 4.3440 & 13.8280 & 4.8980 & 12.7450 \\
        sweep & \texttt{transfuser} & 0 & 0.9118 & +0.47 & +0.07 & 4.6051 & 0.5282 & 0.6151 & 0.8865 & 0.4750 & 3.7640 & 4.3710 & 12.8275 & 4.5030 & 12.0780 \\
        sweep & \texttt{transfuser} & 0.25 & 0.9153 & +0.29 & +1.52 & 4.8027 & 0.3722 & 0.3703 & 0.4809 & 0.5427 & 3.3120 & 3.5900 & 11.8520 & 4.4420 & 11.9100 \\
        sweep & \texttt{transfuser} & 0.5 & 0.9158 & +0.20 & +1.59 & 4.8546 & 0.3340 & 0.2893 & 0.4664 & 0.4407 & 3.2710 & 3.5230 & 11.7300 & 4.4420 & 11.9550 \\
        sweep & \texttt{transfuser} & 0.75 & 0.9112 & +0.14 & +1.88 & 4.8675 & 0.3081 & 0.2427 & 0.4518 & 0.3705 & 3.2220 & 3.4175 & 11.6560 & 4.3890 & 11.8780 \\
        sweep & \texttt{transfuser} & 1 & 0.9004 & +0.11 & +2.03 & 4.8629 & 0.2834 & 0.2142 & 0.4582 & 0.3248 & 3.1855 & 3.3540 & 11.3445 & 4.3690 & 11.8400 \\
        sweep & \texttt{transfuser} & T & 0.8400 & +0.06 & -1.29 & 4.5679 & 0.2320 & 0.2474 & 0.4006 & 0.2626 & 3.8750 & 4.5360 & 13.9250 & 4.8690 & 12.7710 \\
        \bottomrule
    \end{tabular}}
    \caption{Full v1 \textsc{navtest} results for per-teacher
    $\alpha$ sweeps, aggregated across eight shards with
    weights proportional to the number of scenarios.
    The human serves as the SAS reference ($H=P=0$),
    while the base is the shared policy at $\alpha=0$;
    $\alpha=\mathrm{T}$ denotes the native teacher, whose PDMS is taken from \cref{tab:v1}.
    $H$ and $P$ are the SAS axes defined in
    \cref{sec:exp:sas}. Sub-metrics: $\bar v$ mean speed;
    $a^{+}/a^{-}$ peak acceleration/deceleration;
    $j_{\ell}/j_t$ longitudinal/lateral jerk;
    thw/ttc minimum headway/time-to-collision;
    lead/$d_{\min}/d_{\mathrm{vru}}$ minimum
    lead/object/VRU distances.}
    \label{tab:v1full_single}
\end{table*}

\begin{table*}[t]
    \centering
    \scriptsize
    \setlength{\tabcolsep}{3pt}
    \resizebox{\textwidth}{!}{%
    \begin{tabular}{ll c c cc c cccc ccccc}
        \toprule
        scope & teacher & $\alpha$ & PDMS$\uparrow$ & $H$ & $P$ & $\bar v$ & $a^{+}$ & $a^{-}$ & $j_{\ell}$ & $j_{t}$ & thw & ttc & lead & $d_{\min}$ & $d_{\mathrm{vru}}$ \\
        \midrule
        combo & \texttt{ddv2+goalflow 50/50} & 0 & 0.9118 & +0.47 & +0.07 & 4.6051 & 0.5282 & 0.6151 & 0.8865 & 0.4750 & 3.7640 & 4.3710 & 12.8275 & 4.5030 & 12.0780 \\
        combo & \texttt{ddv2+goalflow 50/50} & 0.5 & 0.9520 & +1.05 & +1.59 & 4.8950 & 1.0565 & 0.9788 & 1.7424 & 0.7410 & 3.3390 & 3.7605 & 12.0920 & 4.4030 & 11.6645 \\
        combo & \texttt{ddv2+goalflow 50/50} & 1 & 0.9500 & +2.62 & +2.29 & 4.8309 & 2.4876 & 1.8331 & 3.6637 & 1.7195 & 3.0040 & 3.4865 & 12.5640 & 4.3570 & 11.6665 \\
        combo & \texttt{ddv2+transfuser 70/30} & 0 & 0.9118 & +0.47 & +0.07 & 4.6051 & 0.5282 & 0.6151 & 0.8865 & 0.4750 & 3.7640 & 4.3710 & 12.8275 & 4.5030 & 12.0780 \\
        combo & \texttt{ddv2+transfuser 70/30} & 0.5 & 0.9437 & +1.05 & +1.60 & 4.7260 & 1.0065 & 0.9937 & 1.6054 & 0.8264 & 3.3110 & 3.9040 & 12.8625 & 4.2290 & 11.8385 \\
        combo & \texttt{ddv2+transfuser 70/30} & 1 & 0.9480 & +2.89 & +2.54 & 4.8494 & 2.5248 & 2.0674 & 3.7133 & 2.1054 & 3.1130 & 3.3615 & 12.1140 & 4.1460 & 11.7070 \\
        combo & \texttt{ddv2+transfuser 50/50} & 0 & 0.9118 & +0.47 & +0.07 & 4.6051 & 0.5282 & 0.6151 & 0.8865 & 0.4750 & 3.7640 & 4.3710 & 12.8275 & 4.5030 & 12.0780 \\
        combo & \texttt{ddv2+transfuser 50/50} & 0.5 & 0.9456 & +0.71 & +1.50 & 4.8358 & 0.7557 & 0.6825 & 1.1434 & 0.6745 & 3.2300 & 3.8975 & 12.5060 & 4.4110 & 11.7720 \\
        combo & \texttt{ddv2+transfuser 50/50} & 1 & 0.9475 & +2.09 & +1.78 & 4.9691 & 2.0326 & 1.5691 & 2.8817 & 1.3920 & 3.0725 & 3.6615 & 11.9565 & 4.4330 & 11.8725 \\
        combo & \texttt{ddv2+transfuser 30/70} & 0 & 0.9118 & +0.47 & +0.07 & 4.6051 & 0.5282 & 0.6151 & 0.8865 & 0.4750 & 3.7640 & 4.3710 & 12.8275 & 4.5030 & 12.0780 \\
        combo & \texttt{ddv2+transfuser 30/70} & 0.5 & 0.9089 & +0.46 & +1.94 & 4.7204 & 0.5644 & 0.4788 & 0.7054 & 0.5833 & 3.1725 & 3.4600 & 11.2675 & 4.3330 & 11.9235 \\
        combo & \texttt{ddv2+transfuser 30/70} & 1 & 0.9008 & +0.73 & +2.52 & 4.9728 & 0.7749 & 0.5997 & 1.1111 & 0.7917 & 3.0730 & 3.4105 & 11.2915 & 4.2370 & 11.5775 \\
        combo & \texttt{transfuser+goalflow 50/50} & 0 & 0.9118 & +0.47 & +0.07 & 4.6051 & 0.5282 & 0.6151 & 0.8865 & 0.4750 & 3.7640 & 4.3710 & 12.8275 & 4.5030 & 12.0780 \\
        combo & \texttt{transfuser+goalflow 50/50} & 0.5 & 0.9248 & +0.55 & +1.73 & 4.6543 & 0.5435 & 0.6262 & 0.8534 & 0.6315 & 3.3145 & 3.8360 & 11.8075 & 4.2920 & 11.6380 \\
        combo & \texttt{transfuser+goalflow 50/50} & 1 & 0.9206 & +1.05 & +2.12 & 4.7809 & 1.0321 & 0.9105 & 1.5680 & 0.8838 & 3.2060 & 3.6705 & 11.0475 & 4.2370 & 11.5960 \\
        combo & \texttt{ddv2+gtrs 60/40} & 0 & 0.9118 & +0.47 & +0.07 & 4.6051 & 0.5282 & 0.6151 & 0.8865 & 0.4750 & 3.7640 & 4.3710 & 12.8275 & 4.5030 & 12.0780 \\
        combo & \texttt{ddv2+gtrs 60/40} & 0.5 & 0.9521 & +0.91 & +1.72 & 4.8853 & 0.8642 & 0.8730 & 1.4049 & 0.7874 & 3.1620 & 3.5695 & 11.8470 & 4.4390 & 11.9000 \\
        combo & \texttt{ddv2+gtrs 60/40} & 1 & 0.9543 & +2.66 & +2.35 & 5.0393 & 2.4151 & 1.9628 & 3.5283 & 1.8284 & 2.9325 & 3.2810 & 11.6675 & 4.3950 & 11.8510 \\
        combo & \texttt{ddv2+hydra 60/40} & 0 & 0.9118 & +0.47 & +0.07 & 4.6051 & 0.5282 & 0.6151 & 0.8865 & 0.4750 & 3.7640 & 4.3710 & 12.8275 & 4.5030 & 12.0780 \\
        combo & \texttt{ddv2+hydra 60/40} & 0.5 & 0.9499 & +0.91 & +1.84 & 4.8883 & 0.8822 & 0.8653 & 1.4252 & 0.7692 & 3.1735 & 3.5015 & 11.7565 & 4.4490 & 11.8000 \\
        combo & \texttt{ddv2+hydra 60/40} & 1 & 0.9519 & +2.66 & +2.49 & 5.0141 & 2.3923 & 1.9392 & 3.4845 & 1.8634 & 2.8890 & 3.2330 & 11.8015 & 4.3970 & 11.7950 \\
        combo & \texttt{ddv2+goalflow 60/40} & 0 & 0.9118 & +0.47 & +0.07 & 4.6051 & 0.5282 & 0.6151 & 0.8865 & 0.4750 & 3.7640 & 4.3710 & 12.8275 & 4.5030 & 12.0780 \\
        combo & \texttt{ddv2+goalflow 60/40} & 0.5 & 0.9452 & +1.23 & +1.61 & 4.8364 & 1.1836 & 1.1556 & 1.8826 & 0.8407 & 3.1810 & 3.5640 & 11.9270 & 4.4530 & 11.9990 \\
        combo & \texttt{ddv2+goalflow 60/40} & 1 & 0.9494 & +2.94 & +2.16 & 4.9343 & 2.6060 & 2.0955 & 3.9844 & 2.0162 & 3.0575 & 3.3890 & 11.8915 & 4.4170 & 11.7530 \\
        combo & \texttt{ddv2+goalflow+hydra 40/40/20} & 0 & 0.9118 & +0.47 & +0.07 & 4.6051 & 0.5282 & 0.6151 & 0.8865 & 0.4750 & 3.7640 & 4.3710 & 12.8275 & 4.5030 & 12.0780 \\
        combo & \texttt{ddv2+goalflow+hydra 40/40/20} & 0.5 & 0.9387 & +0.91 & +1.88 & 4.9062 & 0.8805 & 0.8821 & 1.4372 & 0.7460 & 3.1440 & 3.5290 & 11.6810 & 4.4320 & 11.7880 \\
        combo & \texttt{ddv2+goalflow+hydra 40/40/20} & 1 & 0.9496 & +2.40 & +2.27 & 5.0126 & 2.3134 & 1.7531 & 3.3477 & 1.5544 & 2.9870 & 3.3260 & 11.6890 & 4.4140 & 11.7960 \\
        combo & \texttt{ddv2+goalflow+wote 45/45/10} & 0 & 0.9118 & +0.47 & +0.07 & 4.6051 & 0.5282 & 0.6151 & 0.8865 & 0.4750 & 3.7640 & 4.3710 & 12.8275 & 4.5030 & 12.0780 \\
        combo & \texttt{ddv2+goalflow+wote 45/45/10} & 0.5 & 0.9319 & +1.03 & +1.86 & 4.8539 & 0.9849 & 0.9920 & 1.6205 & 0.7741 & 3.1700 & 3.5400 & 11.6570 & 4.4300 & 11.7750 \\
        combo & \texttt{ddv2+goalflow+wote 45/45/10} & 1 & 0.9490 & +2.58 & +2.31 & 4.9875 & 2.3998 & 1.8981 & 3.5852 & 1.6949 & 2.9790 & 3.2710 & 11.9215 & 4.4540 & 11.7420 \\
        combo & \texttt{ddv2+goalflow+gtrs 40/40/20} & 0 & 0.9118 & +0.47 & +0.07 & 4.6051 & 0.5282 & 0.6151 & 0.8865 & 0.4750 & 3.7640 & 4.3710 & 12.8275 & 4.5030 & 12.0780 \\
        combo & \texttt{ddv2+goalflow+gtrs 40/40/20} & 0.5 & 0.9465 & +0.90 & +1.87 & 4.9022 & 0.8959 & 0.8772 & 1.4354 & 0.7268 & 3.1600 & 3.5270 & 11.7180 & 4.4330 & 11.7760 \\
        combo & \texttt{ddv2+goalflow+gtrs 40/40/20} & 1 & 0.9459 & +2.43 & +2.32 & 5.0072 & 2.3323 & 1.7972 & 3.4049 & 1.5534 & 2.9580 & 3.2970 & 11.5530 & 4.4290 & 11.7800 \\
        combo & \texttt{ddv2+goalflow+transfuser+gtrs 35/35/15/15} & 0 & 0.9118 & +0.47 & +0.07 & 4.6051 & 0.5282 & 0.6151 & 0.8865 & 0.4750 & 3.7640 & 4.3710 & 12.8275 & 4.5030 & 12.0780 \\
        combo & \texttt{ddv2+goalflow+transfuser+gtrs 35/35/15/15} & 0.5 & 0.9377 & +0.81 & +1.87 & 4.8538 & 0.7870 & 0.8131 & 1.2800 & 0.7112 & 3.1810 & 3.4970 & 11.6360 & 4.4230 & 11.7960 \\
        combo & \texttt{ddv2+goalflow+transfuser+gtrs 35/35/15/15} & 1 & 0.9402 & +2.11 & +2.33 & 5.0196 & 2.0699 & 1.5952 & 3.0159 & 1.3398 & 3.0240 & 3.3060 & 11.4245 & 4.4000 & 11.7130 \\
        combo & \texttt{goalflow+gtrs 50/50} & 0 & 0.9118 & +0.47 & +0.07 & 4.6051 & 0.5282 & 0.6151 & 0.8865 & 0.4750 & 3.7640 & 4.3710 & 12.8275 & 4.5030 & 12.0780 \\
        combo & \texttt{goalflow+gtrs 50/50} & 0.5 & 0.9246 & +0.52 & +1.82 & 4.8540 & 0.5334 & 0.5611 & 0.8342 & 0.6375 & 3.2320 & 3.5480 & 11.5610 & 4.4120 & 11.7490 \\
        combo & \texttt{goalflow+gtrs 50/50} & 1 & 0.9299 & +1.04 & +2.20 & 4.9193 & 0.9743 & 0.9576 & 1.6231 & 0.8446 & 3.1040 & 3.4010 & 11.2400 & 4.3550 & 11.7250 \\
        \bottomrule
    \end{tabular}}
    \caption{Full v1 \textsc{navtest} results for style blends.
    Each configuration specifies the teacher task vectors and their
    mixing weights, with $\alpha$ controlling the overall strength.
    We report PDMS, SAS axes $H$ and $P$, and the individual
    kinematic and proximity measurements, following the same
    definitions as \cref{tab:v1full_single}. All configurations
    recover the shared base at $\alpha{=}0$.}

    \label{tab:v1full_combo}

    \par\vspace{14pt}
    \noindent\makebox[\textwidth][l]{%
        \begin{minipage}{0.48\textwidth}
            \centering
            \footnotesize
            \setlength{\tabcolsep}{6pt}
            \begin{tabular}{@{}lcc@{}}
                \toprule
                style & discrete codebook & flow (ours) \\
                \midrule
                base ($\alpha{=}0$) & 0.5 & 0.5 \\
                \texttt{ddv2}       & 0.8 & 3.2 \\
                \texttt{goalflow}   & 0.7 & 2.0 \\
                \texttt{wote}       & 0.1 & 0.2 \\
                \texttt{transfuser} & 0.1 & 0.1 \\
                \texttt{hydra\_mdp} & 0.0 & 0.1 \\
                \texttt{gtrs}       & 0.0 & 0.1 \\
                \midrule
                $\Delta H$ & 0.8 & 3.2 \\
                \bottomrule
            \end{tabular}

            \captionof{table}{Per-style assertiveness $H$
            for the decoder ablation in \cref{tab:ablate_dec}.
            Both decoders use the same style conditioning.
            The base is evaluated at $\alpha{=}0$ and
            each style at $\alpha{=}1$.
            The bottom row reports the measured
            assertiveness span $\Delta H$.}
            \label{tab:suppl_ablate_dec}
        \end{minipage}%
    }

\end{table*}

\section{Per-style Ablation Breakdown}
\label{sec:suppl:ablate}
\cref{tab:suppl_ablate_dec} reports the per-style assertiveness scores underlying the decoder comparison in \cref{tab:ablate_dec}. The style span $\Delta H$ is the difference between the largest and smallest measured $H$ values across the evaluated style settings. We include the base case ($\alpha{=}0$) and each style at full strength ($\alpha{=}1$), along with the resulting span.

Flow matching spans a wider range of assertiveness than the discrete codebook ($\Delta H{=}3.2$ versus $0.8$). Most of this gap comes from the more assertive styles: DiffusionDriveV2 reaches $H{=}3.2$ with flow matching but only $0.8$ with the codebook, while GoalFlow reaches $2.0$ and $0.7$, respectively. The smoother styles stay close to the human reference under both decoders. Thus, the wider range in \cref{tab:ablate_dec} mainly reflects the decoder's ability to capture stronger teacher-induced dynamics, rather than a similar increase across all styles.

\end{document}